\documentclass[letterpaper, 10pt, conference]{ieeeconf}      
\IEEEoverridecommandlockouts

\makeatletter
\let\NAT@parse\undefined
\makeatother

\usepackage{cite}
\usepackage{amsmath,amssymb,amsfonts}
\usepackage{algorithmic}
\usepackage{graphicx}
\usepackage{xspace}
\usepackage{stfloats}
\usepackage[table,xcdraw]{xcolor}
\usepackage{booktabs}
\usepackage{makecell}
\usepackage{pifont}
\usepackage{multirow}
\usepackage{caption}
\usepackage{subcaption}
\usepackage{hyperref}
\usepackage{wasysym}

\newcommand{\ours}[0]{HARPER\xspace}
\newcommand{\nsub}[0]{17\xspace}
\newcommand{\nact}[0]{15\xspace}

\newcommand{\mocap}[0]{MoCap\xspace}

\newcommand{\degree}[0]{$^{\circ}$\xspace}
\newcommand{\spot}[0]{Spot\xspace}
\newcommand{\optitrack}[0]{OptiTrack\xspace}

\newcommand{\cc}{\color[rgb]{0,0.6,0.3}\checkmark}
\newcommand{\xx}{\color[rgb]{0.6,0,0}{\ding{55}}}

\def\BibTeX{{\rm B\kern-.05em{\sc i\kern-.025em b}\kern-.08em
    T\kern-.1667em\lower.7ex\hbox{E}\kern-.125emX}}

\makeatletter
\DeclareRobustCommand\onedot{\futurelet\@let@token\@onedot}
\def\@onedot{\ifx\@let@token.\else.\null\fi\xspace}

\def\eg{\emph{e.g}\onedot} 
\def\ie{\emph{i.e}\onedot}

\makeatother

\title{\LARGE \bf Exploring 3D Human Pose Estimation and Forecasting from the Robot's Perspective: The \ours Dataset}

\author{Andrea Avogaro$^{1*}$, Andrea Toaiari$^{1*}$, Federico Cunico$^{1*}$, Xiangmin Xu$^{2}$, Haralambos Dafas$^{2}$, \\ Alessandro Vinciarelli$^{2}$, Emma Li$^{2}$ and Marco Cristani$^{1}$
\thanks{*Equal contribution}%
\thanks{$^{1}$University of Verona, Department of Engineering for Innovation Medicine, Italy. (e-mail: {\tt\small name.surname@univr.it})}%
\thanks{$^{2}$University of Glasgow, School of Computing Science, UK}%
\thanks{Project page: \url{https://intelligolabs.github.io/HARPER}}
}

\begin{document}

\maketitle

\begin{abstract}
We introduce \ours, a novel dataset for 3D body pose estimation and forecast in dyadic interactions between users and \spot, the quadruped robot manufactured by Boston Dynamics. The key-novelty is the focus on the robot's perspective, \ie, on the data captured by the robot's sensors.
These make 3D body pose analysis challenging because being close to the ground captures humans only partially.
The scenario underlying \ours includes \nact actions, of which 10 involve physical contact between the robot and users. The Corpus contains not only the recordings of the built-in stereo cameras of \spot, but also those of a 6-camera \optitrack system (all recordings are synchronized). This leads to ground-truth skeletal representations with a precision lower than a millimeter. In addition, the Corpus includes reproducible benchmarks on 3D Human Pose Estimation, Human Pose Forecasting, and Collision Prediction, all based on publicly available baseline approaches. This enables future \ours users to rigorously compare their results with those we provide in this work.
\end{abstract}
\section{Introduction}
\begin{figure}[!tb]
    \centering
    \includegraphics[width=\linewidth]
    {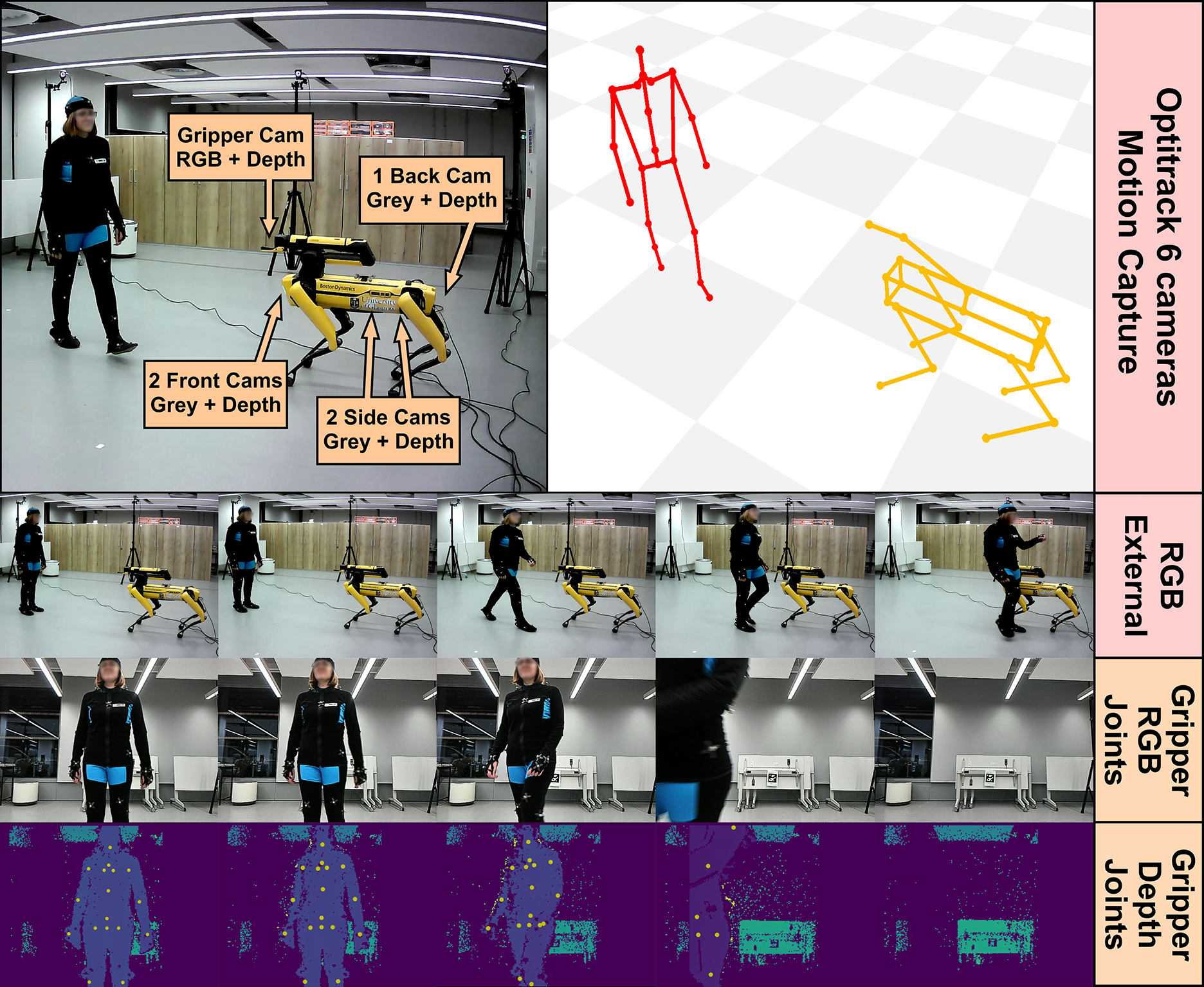}
    \caption{
    \ours Showcase. (Top-left) We exploit the \spot onboard equipment to let the robot perceive people. (Top-right) thanks to a 6-camera \optitrack setup we provide 3D human poses represented with 21-joints with $0.035$ mm of error, used as reference. (Second row) an additional external RGB camera shows the actions performed. (Third row) The Gripper cam RGB Point of View: the yellow dots are the joints back-projected in the image plane. (Fourth row). The Gripper cam Depth Point of View, with the ground truth joints. Zoom the figure for a better view of the joints.
    }
    \label{fig:teaser} 
\end{figure}

One of the main changes characterizing the transition from Industry 4.0 to Industry 5.0 is the shift from Human-Robot \emph{Interaction} to Human-Robot \emph{Collaboration}~\cite{maddikunta2022industry}. This shift necessitates the evolution of robots into cobots, that is, intelligent platforms equipped with capabilities like visual perception, action recognition, intent prediction, and safe online motion planning. These technologies empower cobots with human awareness, enabling them to adapt their behavior in real-time, which is a stark contrast to the rigid, pre-programmed routines of traditional cobots~\cite{ElZaatari2019}. In other words, making sense of human behavior is a key-requirement for a robot to become a cobot and, correspondingly, to be capable of adaptive and seamless interaction with its users~\cite{wang2021deep}.

Thus motivated, 
we propose \emph{Human from an Articulated Robot Perspective} (\ours), a new, publicly available dataset revolving around the interaction between human users and \spot, the quadruped robot manufactured by Boston Dynamics. Such a platform attracts increasingly more attention and, not surprisingly, it was recently included in \emph{Habitat 3.0}~\cite{puig2023habitat}, one of the most popular environments for simulating Human-Robot interactions. In addition, \spot is an ideal cobot candidate for at least three reasons: the first is that the four-leg design and the biologically-inspired locomotion provide the 
ability to operate on diverse and challenging terrains (the robot can even climb stairs~\cite{biswal2021development}), thus making of \spot a potential 
companion in a wide spectrum of settings~\cite{merkt2019towards,guertler2023robot,halder2023construction}. The second is that \spot is equipped with one of the most advanced self-balancing systems available in the market and this significantly limits the risk of accidents in physically close interaction with the users. 
The third is that 
\spot 
is equipped with a total of 5 \mbox{greyscale + depth} sensors mounted on its body and an \mbox{RGB-D} camera on its grasper arm 
(see Fig.~\ref{fig:teaser}). 
This is important because such a sensing apparatus makes \spot particularly suitable for analysis and understanding of human behavior, a key-step in the evolution from robot to cobot (see above). 

\ours includes dyadic interactions between \spot and 17 human users, 5 females and 12 males, each performing \nact actions that require different degrees of collaboration with the robot (see Section~\ref{data} for more details). The data captured with the \spot sensors (see above) were enriched with the recordings of a 6-camera \optitrack motion capture (MoCap) system capable of extracting skeletal models of the users. The joints were localized with a precision of less than one millimeter (see Figure~\ref{fig:teaser}), thus providing highly accurate ground-truth information about the pose and position of the users. This is a major advantage because \spot sensors and \mocap cameras are synchronized. Therefore, skeletal models can be used to reliably validate approaches for human behavior analysis and understanding based on the sole \spot sensors. 

In addition to the above, skeletal representations enable one of the key-novelties of \ours, namely the possibility to train approaches capable of recognizing 3D body pose and movement when the \spot, due to its limited height, can ``see'' its users only partially, something that happens whenever the distance is small. To the best of our knowledge, this is one of the first datasets that allows the investigation of such a problem in 3D. 

We asked the 17 \ours participants to stage two major types of physical contact with the robot, namely \emph{unintentional} and \emph{intentional}, according to the terminology proposed in~\cite{contact2020IEEERAS}. The first type includes (staged) collisions, while the second includes punches, kicks and soft touches. We paid special attention to the first type because of the major role collisions play in scenarios based on co-located interactions. Correspondingly, 

we enriched \ours with benchmarks, \ie, reproducible experimental protocols and baseline approaches designed to address three tasks relevant to the analysis of physical contact, namely 3D Human Pose Estimation (especially when \spot can ``see'' its users only partially), 3D Human Pose Forecasting and Collision Prediction. This allows researchers interested in \ours to rigorously compare their results with those presented in this article (see Section~\ref{evaluation}).

Overall, the main contributions of the paper can be summarized as follows:
\begin{itemize}
\item We propose the first dataset that includes not only the ``point of view'' of the robot (the data captured with the sensors of the \spot), but also a panoptic point of view (the data captured with the \mocap system) that provides accurate ground-truth information for position and pose of both users and robot;
\item To the best of our knowledge, \ours is the first dataset enabling the reconstruction of the human users' pose with the data captured with a quadruped robot, a problem which is challenging because \spot is small (hence, the cameras cannot capture the whole body of the user);
\item \ours allows, for the first time, visual prediction of collisions between a mobile robotic platform and users.
\end{itemize}
The rest of this paper is organized as follows: Section~\ref{survey} surveys previous work, Section~\ref{data} describes \ours in detail, Section~\ref{evaluation} presents the benchmarks, and the final Section~\ref{concl} draws some conclusions.

\section{Related Work}\label{survey}

\begin{table*}[t]
    \caption{
    Main HRI datasets revolving around human movement and its analysis. 
    Values in the participants column indicated with the asterisk (*) refer to datasets captured in uncontrolled scenarios.
    }
    \begin{center}
\resizebox{\textwidth}{!}{%
        \begin{tabular}{c|cc|ccccccc}
        \toprule
        Dataset & Participants & Actions & 
        \begin{tabular}[c]{@{}c@{}}Mobile\\ Robot\end{tabular} &
        \begin{tabular}[c]{@{}c@{}}Robot\\ POV\end{tabular} & 
        \begin{tabular}[c]{@{}c@{}}Human\\ Skeleton\end{tabular} & 
        \begin{tabular}[c]{@{}c@{}}Human\\ Joints\end{tabular} &  
        \begin{tabular}[c]{@{}c@{}}Marker-Based\\ \mocap\end{tabular} &  
        \begin{tabular}[c]{@{}c@{}}Robot\\ Skeleton\end{tabular} & 
        \begin{tabular}[c]{@{}c@{}}Collisions\\/ Intended Contact\end{tabular} \\
        \midrule
        THÖR~\cite{thor} & 9 & 13 & \cc & \xx & \xx & \xx & \cc & \xx & \xx \\
        THÖR-Magni~\cite{schreiter2022magni} & 40 & 5 & \cc & \cc & \xx & \xx & \cc & \xx & \xx \\
        JRDB~\cite{martin2021jrdb} & 3.5K* & N/A & \cc & \cc & \xx & \xx & \xx & \xx & \xx \\
        L-CAS Multisensor~\cite{yan2018multisensor} & N/A* & N/A & \cc & \cc & \xx & \xx & \xx & \xx & \xx \\
        FROG~\cite{amodeo2023frog} & 1M* & N/A & \cc & \cc & \xx & \xx & \xx & \xx & \xx \\
        CODa~\cite{zhang2023towards} & N/A* & N/A & \cc & \cc & \xx & \xx & \xx & \xx & \xx \\
        PTUA~\cite{zhang2023robot} & N/A & N/A & \cc & \cc & \xx & \xx & \cc & \xx & \xx \\
        InHARD~\cite{dallel2020inhard} & 16 & 14 & \xx & \xx & 3D & 17 & \xx & \xx & \xx \\
        JRDB-Pose~\cite{vendrow2023jrdb} & 5K* & N/A & \cc & \cc & 2D & 17 & \xx & \xx & \xx \\
        HuRoN~\cite{hirose2023sacson} & N/A* (5/17 for exp) & N/A & \cc & \cc & \xx & \xx & \xx & \xx & \cc \\
        NatSGD~\cite{shrestha2023natsgd} & 18 & 11 & \xx & \cc & estim. 2D & 25 & \xx & Arm & \xx \\
        CHICO~\cite{sampieri2022pose} & 20 & 7 & \xx & \xx & 2D, 3D & 15 & \xx & Arm & \cc \\
        SCAND~\cite{karnan2022socially} & N/A* (14 for exp) & 12 & \cc & \cc & \xx & \xx   &\xx & Quadruped, Wheeled & \xx \\
        UF-Retail-HRI~\cite{chen2022human} & 8 & 2 & \cc & \cc & 3D & 23 & \xx & Arm & \xx \\
        \midrule
        \textbf{\ours} & \nsub & \nact & \cc & \cc & 2D, 3D & 21 & \cc & Quadruped & \cc \\
        \bottomrule
        \end{tabular}%
}
        \label{tab:datasets}
    \end{center}
\end{table*}

Table~\ref{tab:datasets} shows the main differences between \ours and existing datasets of similar scope. Most available corpora are based on the analysis of people's trajectories. The THÖR dataset~\cite{thor}, a well-known example, contains the 2D trajectories of 9 human users moving together with a robot. Besides this, the data includes 6D head positions, LiDAR data from a stationary sensor, orientations and eye gaze direction for the participants. 
THÖR-Magni~\cite{schreiter2022magni}, the second version of THÖR, introduces onboard sensors on the mobile robot and semantic attributes describing the roles and activities of detected people. In a similar vein, the JRDB~\cite{martin2021jrdb} aims at enabling mobile robots to detect and track humans in both indoor and outdoor settings. The data includes stereo cylindrical RGB videos and LiDAR point clouds collected and annotated with 2D and 3D bounding boxes, respectively. In addition, the dataset includes benchmarks for both 2D and 3D detection and tracking. A more recent version of the corpus includes 2D human-pose skeletal annotations~\cite{vendrow2023jrdb}. 

Other datasets provide information about objects that the robots can encounter while moving. For example, CODa~\cite{zhang2023towards} aims at both object detection and semantic segmentation. It was acquired with a wheeled robot, featuring sequences in indoor and outdoor settings on a university campus 3D semantic segmentation and 3D object detection benchmarks. In the case of FROG~\cite{amodeo2023frog}, based on LiDAR sensors placed on a wheeled robot at roughly the height of human knees, the problem is the detection of people in possibly crowded sites where humans can be confused with static and dynamic obstacles. A similar issue is at the core of the dataset proposed in~\cite{zhang2023robot}, where the material is collected with an RGB-D camera mounted on a small mobile robot. The annotations include attributes such as, \eg, the presence of static obstacles, illumination and humans' poses. An \optitrack \mocap system provides information about the position of both the robot and users. The problem of navigating through an environment, possibly shared with humans, is the focus of HuRon~\cite{hirose2023sacson}. The data was collected with a Roomba bot equipped with LiDAR, bumper collision detectors, video and odometry sensors. However, no pose annotation is provided about the people sharing the space with the robot.

\ours shows major novelties with respect to the datasets above. The availability of 3D skeletons for both the robot and users provides unprecedentedly detailed information about the interaction between the two, especially when taking into account that the joints are localized with a precision of less than one millimeter.
A similar acquisition precision is achieved with InHARD~\cite{dallel2020inhard}, an industrial HRI dataset featuring both RGB images and \mocap data of a person performing multiple manual tasks, captured with wearable devices. A robotic arm, mostly stationary, is the platform used for the experiments and it never collides with the user, offering a looser type of interaction. This is not the case in \ours which includes physical contacts of different types. In~\cite{chen2022human}, a mobile wheeled robot is employed to capture an HRI dataset in a retail environment. Multiple people navigate the room and perform picking and sorting actions while the robot moves with them. Egocentric videos, scene videos, eye gaze directions, point clouds, and other data are collected. The human poses are collected through an IMU-based \mocap device, which requires careful setup and calibration for every person. However, the \spot used in \ours is a more advanced robotic platform, and its movement is significantly less constrained.

Skeleton representations were used in other corpora too. In~\cite{sampieri2022pose}, the scenario is a collaboration between a user and a robotic arm in an industrial setting. A \mocap system captures the skeleton of the user from an external point of view, missing the robot's perspective (unlike \ours). Furthermore, the acquisition is markerless and, therefore, the joint localization is less precise. 
In another dataset, the multimodal NatSGD~\cite{shrestha2023natsgd}, the goal is imitation learning, and the data includes human commands, such as speech and gestures, with a focus on robot behaviour in the form of synchronized demonstrated robot trajectories. However, the joint localization is, once again, less precise than in \ours because it is performed by applying Openpose to videos.
Finally, to the best of our knowledge, the only other dataset in which the robot \spot was actually involved is SCAND~\cite{karnan2022socially}, where two robots, a wheeled one and the \spot, are teleoperated in human-populated environments. A large variety of data is acquired thanks to an additional LiDAR sensor mounted on the two robots. However, no skeletal models are considered for humans, a major difference with respect to \ours. The dataset we propose appears to have distinctive characteristics with respect to those currently available in the literature.
\section{The \ours Dataset}\label{data}
\label{sec:dataset}
\begin{figure}[!th]
    \centering
    \includegraphics[width=0.8\linewidth]{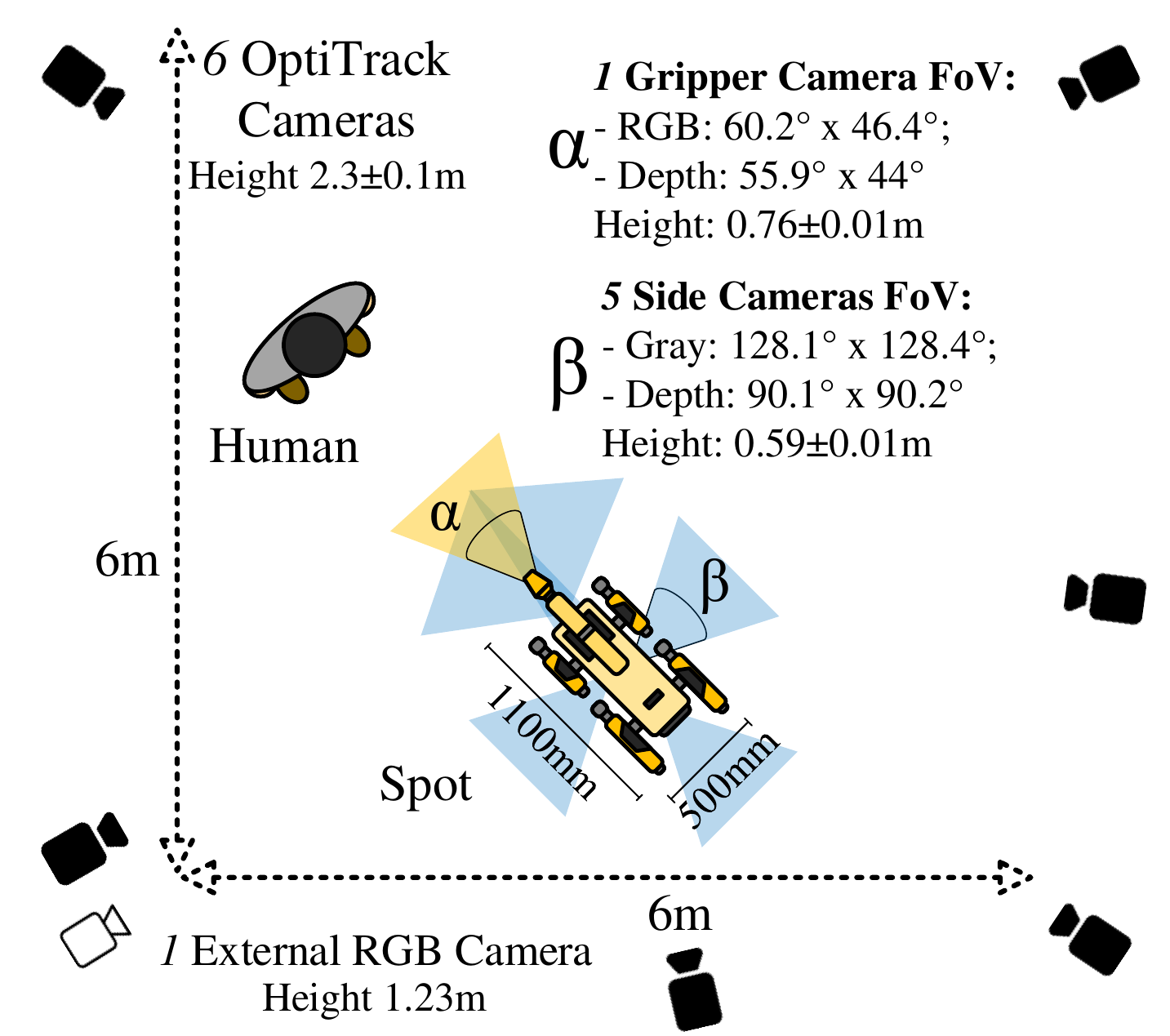}
    \caption{A 6-camera \optitrack system covers a $6\times6$ squared meters area where users and \spot can freely move. The external RGB camera's field of view covers the setting. The 5 \spot on-body greyscale + depth cameras and the RGB-D frontal camera (gripper) cover the environment surrounding the robot.
    }
    \label{fig:lab-map} 
\end{figure}

The main motivation behind the design of \ours is to expand the research opportunities enabled by previous HRI datasets (see Section~\ref{survey}), especially towards the transition from robots to cobots.  The collection of the corpus involved \nsub participants who were asked to perform \nact actions (the same for all participants). The data was captured with the sensors equipped on \spot: 5 greyscale + depth sensors and one RGB-D camera mounted on the gripper. Moreover, we used 6 \mocap sensors (\optitrack) and one RGB camera capturing the full setting (see Fig.~\ref{fig:lab-map}). 
Overall, \ours contains 607 sequences for a total of over 60000 RGB images, grayscale images, depth frames, and 3D data from multi-sensor recordings. 
In the following, we discuss the acquisition setup (Sec.~\ref{subsec:setup}), we describe the actions we captured and their annotations (Sec.~\ref{subsec:procedure}), and, finally,  we provide key-statistics about the data (Sec.~\ref{subsec:data-stats}). 

\subsection{Acquisition Setup}
\label{subsec:setup}

We collected all data in a laboratory (the layout is in Fig.~\ref{fig:lab-map}). The 6 cameras of the \optitrack \mocap system were arranged to cover a $6 \times 6$ $m^2$ area, free of obstacles, in which the participants performed the 15 actions of the \ours scenario. All participants wore a motion capture suit with 37 reflective markers distributed according to the \optitrack \emph{Baseline Marker Set} configuration. After calibration, the \optitrack tracks the markers with a $0.035$ mm error at a sampling frequency of 120Hz. Furthermore, thanks to the configuration above, the \optitrack software (Motive) automatically extracts a 21-joint skeleton representation based on the marker positions.

The robot involved in the experiment is the Boston Dynamics \spot, a 12-DoF (3 per leg) quadruped robot equipped with five stereo cameras (greyscale ones and depth) around its body and one RGB-D camera on the gripper. The \spot acts within the \optitrack area described above, and its skeleton is obtained by applying forward kinematics to its internal motors state, acquired through the API provided by Boston Dynamics. The \spot skeleton is then positioned in the same 3D scene as the participants' skeletons using a 4-marker rigid body mounted on its back and tracked by the \optitrack. 

The framerate of \spot cameras is roughly 10 FPS.
The data captured with the \spot are synchronized with those captured with the \optitrack. This ensures one of the most distinctive features of \ours, namely the availability of two points of view, the one of the robot and a panoptic one that covers the whole scene. The synchronization was obtained by taking into account the timestamps of the data and the temporal alignment error is lower than 2 milliseconds. 
It is worth noticing that the overlap between \spot cameras is limited to the 3 frontal cameras with a very partial overlap.
As a reference, we added an external RGB camera positioned outside the \optitrack delimited area that captured the whole scene (see bottom left part of Fig.~\ref{fig:lab-map}). All the videos recorded with such a camera are provided with the dataset. 

\subsection{Actions and Annotations}
\label{subsec:procedure}

We involved 17 university students as participants in the data collection (5 females and 12 males). They all signed an informed consent letter, and all information they provided, including the data collected during their participation, was treated according to the ethical regulations of the university in which the material was collected. Every participant interacted with the \spot individually in a session that included multiple steps (always the same and always in the same order). 
First, the participants were helped to wear the suit necessary for marker tracking (see above), and then they were asked to display a T-pose for calibrating the skeleton extraction.

After calibrating the \optitrack, we asked the participants to perform 15 actions designed to reproduce different situations (see Table~\ref{tab:actions}), including 8 in which the robot stands still and 7 in which the robot moves. In particular, the participants were instructed to act collisions as realistically as possible, \ie, as if they were accidentally and unintentionally bumping into the Spot.
The area covered by the \optitrack is sufficiently wide to perform the actions comfortably (see above), but some participants still moved inadvertently out of it, thus leading to missed markers in a few frames. Similarly, some occlusions prevented the \optitrack from working properly in a few moments. However, these issues concerned no more than 3\% of the total frames and missing markers were effectively replaced through linear interpolation, thus ensuring that the skeleton representation was acquired with continuity and with the same precision at all times. 

As \optitrack and \spot share the same reference system, it was possible to project the 3D skeletons onto the videos captured with the robot's cameras (greyscale and RGB). In this way, the videos were annotated with the correct positions of all joints. In addition, given that the robot's leg motor state is known, forward kinematics was applied to compute the position of the robot's joints in the 3D space. This allowed us to obtain a 21-joint representation not only of the participants' skeletons but also of the robot's skeleton. 

\begin{table}[t]
    \caption{\ours Actions. The expression \emph{Contact} means that the distance between \spot and user is lower than $10$ cm.}
    \begin{center}
\resizebox{\linewidth}{!}{%
        \begin{tabular}{clcc}
        \toprule
        \textbf{Action} & \textbf{Action Description} & \begin{tabular}[c]{@{}c@{}}\textbf{Robot}\\ \textbf{Moving}\end{tabular} & \textbf{Contact} \\
        \midrule
    
        \begin{tabular}[c]{@{}c@{}}A1\\Walk+Crash\\Frontal\end{tabular}& \begin{tabular}[l]{@{}l@{}}Human walks towards \spot\\oriented frontally then collides;\end{tabular} &  
        & \checkmark 
        \\
        \midrule
        
        \begin{tabular}[c]{@{}c@{}}A2\\Walk+Crash\\45\degree\end{tabular}& \begin{tabular}[l]{@{}l@{}}Human walks towards \spot\\oriented at 45\degree then collides;\end{tabular} &  & \checkmark \\
        \midrule
        
        \begin{tabular}[c]{@{}c@{}}A3\\Walk+Crash\\Sideway\end{tabular}& \begin{tabular}[l]{@{}l@{}}Human walks towards \spot\\oriented at 90\degree then collides;\end{tabular} &  & \checkmark \\
        \midrule
        
        \begin{tabular}[c]{@{}c@{}}A4\\Walk+Crash\\Backwards\end{tabular}& \begin{tabular}[l]{@{}l@{}}Human walks towards \spot\\oriented backwards then collides;\end{tabular} &  & \checkmark \\
        \midrule

        \begin{tabular}[c]{@{}c@{}}A5\\Walk+Stop\end{tabular} & \begin{tabular}[l]{@{}l@{}}Human walks towards \spot,\\then stops right before colliding;\end{tabular} & \checkmark &  \\
        \midrule
        
        \begin{tabular}[c]{@{}c@{}}A6 \& A7\\Walk+Avoid\end{tabular} & \begin{tabular}[l]{@{}l@{}}Human and \spot walk towards each\\other avoiding collision at last second\\on the right (A6) / left (A7).\end{tabular} & \checkmark &  \\
        \midrule

        \begin{tabular}[c]{@{}c@{}}A8\\Walk+Touch\end{tabular} & \begin{tabular}[l]{@{}l@{}}Human walks towards \spot, then\\physically touch it; \end{tabular} &  & \checkmark \\
        \midrule
        \begin{tabular}[c]{@{}c@{}}A9\\Walk+Kick\end{tabular} & \begin{tabular}[l]{@{}l@{}}Human walks towards \spot, then kicks it; \end{tabular} &  & \checkmark \\
        \midrule
        
        \begin{tabular}[c]{@{}c@{}}A10 \& A11\\Walk+Punch\end{tabular} & \begin{tabular}[l]{@{}l@{}}Human walks towards \spot oriented\\at 0\degree (A10) / 90\degree (A11), then punches it\end{tabular} &  & \checkmark \\
        \midrule

        \begin{tabular}[c]{@{}c@{}}A12\\Circular Walk\end{tabular}& \begin{tabular}[l]{@{}l@{}}Human and \spot walk together in a \\circular path\end{tabular} & \checkmark &  \\
        \midrule
        
        \begin{tabular}[c]{@{}c@{}}A13\\Circular Follow\\+Touch\end{tabular}& \begin{tabular}[l]{@{}l@{}}Human follows \spot in a circle,\\then touches it with the hand\end{tabular} & \checkmark & \checkmark \\
        \midrule

        \begin{tabular}[c]{@{}c@{}}A14\\Circular Follow\\+ Avoid\end{tabular} & \begin{tabular}[l]{@{}l@{}}\spot follows the human in a circle,\\then avoids contact\end{tabular} & \checkmark &  \\
        \midrule
        
        \begin{tabular}[c]{@{}c@{}} A15\\Circular Follow\\+ Crash\end{tabular} & \begin{tabular}[l]{@{}l@{}}\spot follows the human in a circle,\\then a collision happen\end{tabular} & \checkmark & \checkmark \\
        
        \bottomrule
        \end{tabular}%
}
    \end{center}
   \label{tab:actions}
\end{table}

\subsection{Dataset Statistics}
\label{subsec:data-stats}

\begin{figure}[!tb]
    \centering
    \includegraphics[width=\linewidth]{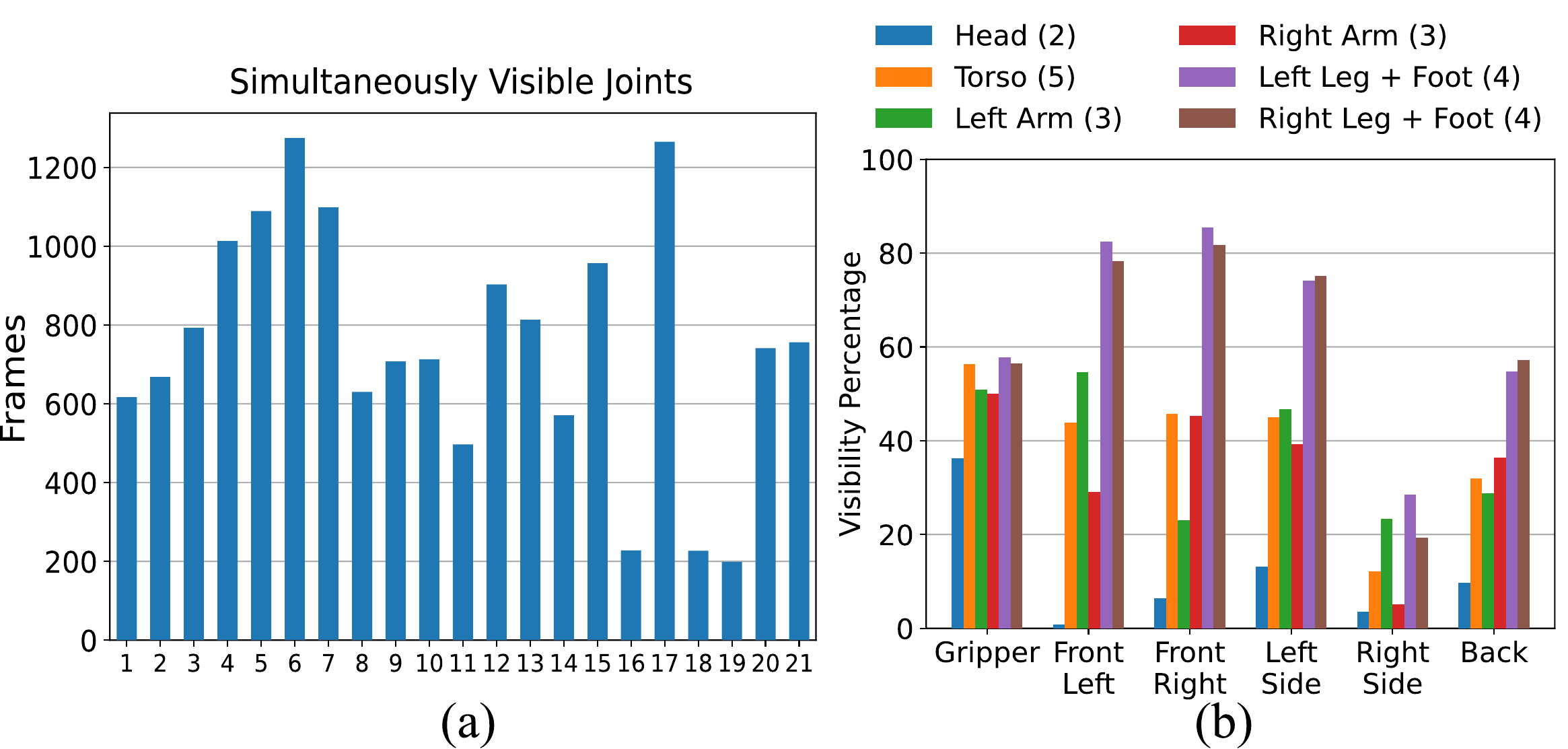}
    \caption{
    Joints visibility from the robot's perspective. The left chart shows how many frames contain exactly $n$ joints for $n=1,\ldots,21$. The right plot shows the percentage of frames in which the different parts of the skeleton are visible.
    }
    \label{fig:stats-visibility}
\end{figure}

\begin{figure}[!tb]
    \centering
    \includegraphics[width=\linewidth]{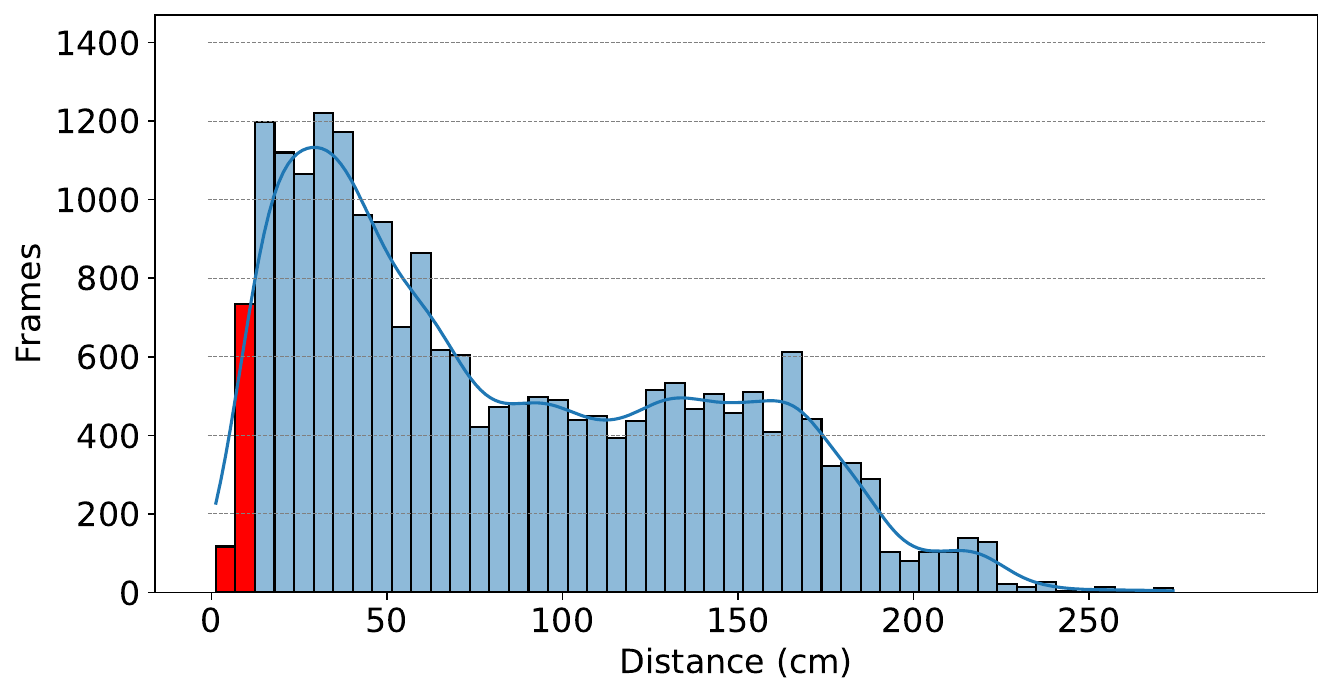}
    \caption{Distribution of distances between \spot and users (the distance considers the two closest joints of human and robot). Red columns correspond to distances lower than $10$ cm, considered as cases of physical contact.
    }
    \label{fig:stats-dist}
\end{figure}

Fig.~\ref{fig:stats-visibility}a shows, for all possible values of $n$, the number of frames in which exactly $n$ human skeleton joints are visible to the robot. Such information is important to understand the level of difficulty in addressing one of the new tasks \ours is enabling, namely analysis and understanding of human pose when this latter is only partially visible. Similar information is shown in Fig.~\ref{fig:stats-visibility}b, where human joints are grouped according to five body parts, \ie head (2 joints), torso (5 joints), left/right arm (3 joints), and left/right leg (4 joints). The figure reports the percentage of times such body parts are visible (one joint is sufficient for the part to be considered visible) to each camera. One of the main patterns is that the gripper camera is more likely to capture the upper part of the body and legs, but not the feet (\ie the spike on 17 visible joints caused by the gripper camera Field of View), while the other on-board cameras are more likely to capture the limbs.

For what concerns the interaction between \spot and the participants,  Fig.~\ref{fig:stats-dist} shows the histograms of the distances between the closest joints of the two. Two modes appear, namely below and above 1.3 meters of distance. Distances corresponding to physical contact are in red. A threshold distance of 10 cm was used to discriminate whether physical contact is happening (see details in Sec.~\ref{sec:collision-detection}).
\section{Experimental Evaluation}\label{evaluation}
\label{sec:evaluation}

\ours provides three benchmarks, one on \textit{3D Human Pose Estimation} (3D-HPE), one on \textit{3D Human Pose Forecasting} (3D-HPF) and one on \textit{Collision Prediction} (CP). All benchmarks are in \emph{robot's perspective}, \ie, they are based on the data captured with the robot's sensors, one of the key-novelties of \ours.

Participants S1-12 were used for training (15984 frames), while participants S13-S17 were used for testing (5542 frames). For 3D-HPF, we sampled 7917 sequences (of 20 frames each) for the training set and 3088 for the test set, keeping the same distribution of participants. The sequences were sampled by using a rolling window with a step of 1 frame. We excluded from the test set the sequences that do not contain any visible joint. 

\subsection{3D Human Pose Estimation}
\label{subsec:pose-estimation-eval}
\begin{figure}[!tb]
    \centering
    \includegraphics[width=\linewidth]{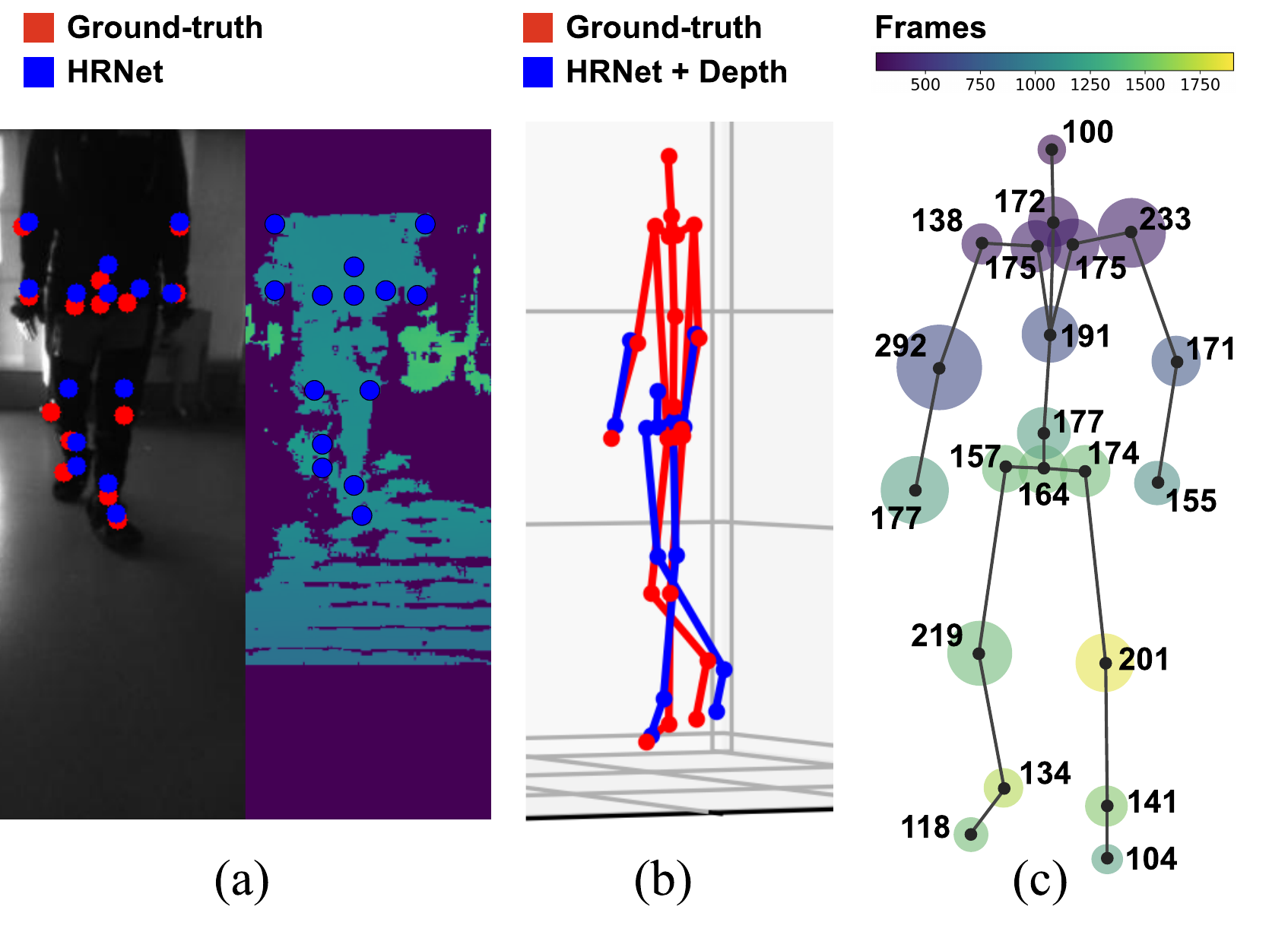}
    \caption{3D human pose estimation from the robot results. (a) On the left, the predicted 2D joints (in blue) by HRNet~\cite{sun2019deep} and the corresponding ground truth joints (in red). On the right, the depth image with the same 2D detections overimposed. The depth will serve to do the lifting. (b) The lifted 3D poses alongside the complete \optitrack skeletons. (c) MPJPE (in mm) for every visible joint (inside the depth FOV) on the test set. The size of the blobs is proportional to the errors, while colors are related to the number of times a joint is visible from the robot's perspective.}
    \label{fig:pose-estimation}
\end{figure}
3D-HPE in robot's perspective is the task of finding the 3D coordinates of the visible human joints when using as input greyscale images and synchronized depth maps captured with the robot's sensors. The main challenge is that humans are not necessarily fully visible (see Section~\ref{data}). Therefore, the proposed baseline approach first uses a 2D pose estimator to find the position of the visible joints and then compute their 3D positions. Such a task is performed by exploiting the depth values as shown in~\cite{liu2022simple} (see Figure~\ref{fig:pose-estimation}). 

The 2D pose estimator is HRNet~\cite{sun2019deep}, trained on \ours training data after resizing the images to $256 \times 256$ (no augmentation is applied). The Field Of View (FOV) of the depth sensors is narrower than the one of the video cameras. Therefore, when the depth value is not available for an estimated joint because out of the depth FOV, it is considered as non-visible. 
The positions of the joints, with their corresponding depth values, can then be mapped into the 3D \optitrack system of coordinates. Once such a task is performed, the 3D points inferred by the approach can be compared with those of the \mocap ground-truth skeleton.

We evaluated 2D pose estimation performance with the Percentage of Correct Keypoints (PCK)~\cite{yang2012articulated}, 
\ie, the fraction of correct predictions within a distance threshold $\tau$ (set to $0.5$ on the predicted heatmaps). For the 3D joints estimation, we used the Mean Per Joint Position Error (MPJPE)~\cite{joo2015panoptic},  \ie, the mean Euclidean distance between the visible estimated joints and the ground-truth \optitrack ones.
 
We obtained a PCK of 82.2\% and an average MPJPE of 168 mm on 2D and 3D poses, respectively (see Fig.~\ref{fig:pose-estimation}). 
The 2D baseline performs well, especially when taking into account that, in many cases, only one limb is visible or the participant is very close to the \spot.

The 3D lifting shows some limitations due to the noise in the depth maps, especially when the participants are far from the \spot. However, the performance was sufficient to address 3D-HPF and CP, both in robot's perspective.

\subsection{3D Human Pose Forecasting}
\label{subsec:pose-forecasting-eval}

\begin{table*}[!ht]\centering
    \setlength{\tabcolsep}{2.5pt} 
    \caption{Pose forecasting errors. We provide the MPJPE expressed in mm with a prediction horizon of 400 and 1000 ms. The errors are computed for the particular frame for each action (first nine columns) as well as the average over all frames (\emph{Average}), and the average over the last frame of each action instance (\emph{Last frame average}). 
    }\label{tab:pose-forecasting}
    \scriptsize
    \resizebox{\textwidth}{!}{%
        \begin{tabular}{cc|cc|cc|cc|cc|cc|cc|cc|cc|cc|cc||cc|cc}\toprule
        \multicolumn{2}{c|}{\textbf{Actions}} & \multicolumn{2}{c}{\textbf{A1-4}} & \multicolumn{2}{c}{\textbf{A5}} &\multicolumn{2}{c}{\textbf{A6-7}} &\multicolumn{2}{c}{\textbf{A8}} &\multicolumn{2}{c}{\textbf{A9}} &\multicolumn{2}{c}{\textbf{A10-11}} &\multicolumn{2}{c}{\textbf{A12}} &\multicolumn{2}{c}{\textbf{A13}} &\multicolumn{2}{c}{\textbf{A14}} &\multicolumn{2}{c||}{\textbf{A15}}&\multicolumn{2}{c}{\textbf{\begin{tabular}[c]{@{}c@{}}\emph{Average}\end{tabular}}} &\multicolumn{2}{c}{\textbf{\begin{tabular}[c]{@{}c@{}}\emph{Last frame}\\\emph{average}\end{tabular}}}  \\ \cmidrule{1-26}
        \multicolumn{2}{c|}{\textbf{Time (ms)}} &\underline{\textit{400}} &\underline{\textit{1000}} &\underline{\textit{400}} &\underline{\textit{1000}} &\underline{\textit{400}} &\underline{\textit{1000}} &\underline{\textit{400}} &\underline{\textit{1000}} &\underline{\textit{400}} &\underline{\textit{1000}} &\underline{\textit{400}} &\underline{\textit{1000}} &\underline{\textit{400}} &\underline{\textit{1000}} &\underline{\textit{400}} &\underline{\textit{1000}} &\underline{\textit{400}} &\underline{\textit{1000}} &\underline{\textit{400}} &\underline{\textit{1000}} &\underline{\textit{400}} &\underline{\textit{1000}}&\underline{\textit{400}} &\underline{\textit{1000}}
        \\\midrule

\multirow{3}{*}{STSGCN} & GT & 127 & 195 & 116 & 167 & 117 & 169 & 112 & 154 & 154 & 237 & 145 & 231 & 158 & 251 & 140 & 224 & 129 & 183 & 170 & 278 & 129 & 197 & 158 & 288 \\ 
 & GT+R & 198 & 249 & 136 & 177 & 391 & 461 & 137 & 162 & 169 & 246 & 164 & 239 & 162 & 242 & 144 & 206 & 306 & 357 & 343 & 389 & 210 & 265 & 234 & 346 \\ 
 & HRNet+D+R & 373 & 416 & 170 & 234 & 529 & 640 & 172 & 184 & 206 & 294 & 221 & 292 & 208 & 290 & 184 & 267 & 484 & 582 & 531 & 581 & 313 & 374 & 332 & 446 \\ 
\midrule
\multirow{3}{*}{SiMLPe} & GT & 62 & 149 & 60 & 141 & 40 & 97 & 30 & 72 & 64 & 143 & 76 & 181 & 90 & 210 & 62 & 149 & 44 & 101 & 93 & 222 & 59 & 140 & 97 & 264 \\ 
 & GT+R & 164 & 246 & 106 & 178 & 366 & 473 & 87 & 122 & 84 & 158 & 113 & 200 & 116 & 225 & 79 & 158 & 262 & 346 & 300 & 373 & 169 & 249 & 204 & 372 \\ 
 & HRNet+D+R & 388 & 475 & 185 & 256 & 674 & 929 & 169 & 207 & 272 & 417 & 368 & 549 & 222 & 337 & 211 & 301 & 518 & 654 & 628 & 769 & 373 & 501 & 441 & 687 \\ 
 \midrule
\multirow{3}{*}{EqMotion} & GT & 43 & 112 & 39 & 91 & 25 & 62 & 23 & 59 & 42 & 103 & 60 & 136 & 68 & 167 & 51 & 122 & 34 & 92 & 75 & 167 & 43 & 104 & 70 & 196 \\ 
 & GT+R & 151 & 217 & 89 & 131 & 344 & 416 & 79 & 107 & 65 & 126 & 101 & 168 & 106 & 198 & 71 & 132 & 257 & 334 & 294 & 352 & 156 & 217 & 182 & 311 \\ 
 & HRNet+D+R & 362 & 439 & 166 & 209 & 526 & 620 & 163 & 198 & 190 & 239 & 240 & 297 & 198 & 290 & 172 & 230 & 478 & 564 & 545 & 568 & 309 & 372 & 333 & 474 \\ 
        \bottomrule
    \end{tabular}%
}
\end{table*}

\begin{figure}[t]
    \centering
    \includegraphics[width=\linewidth]{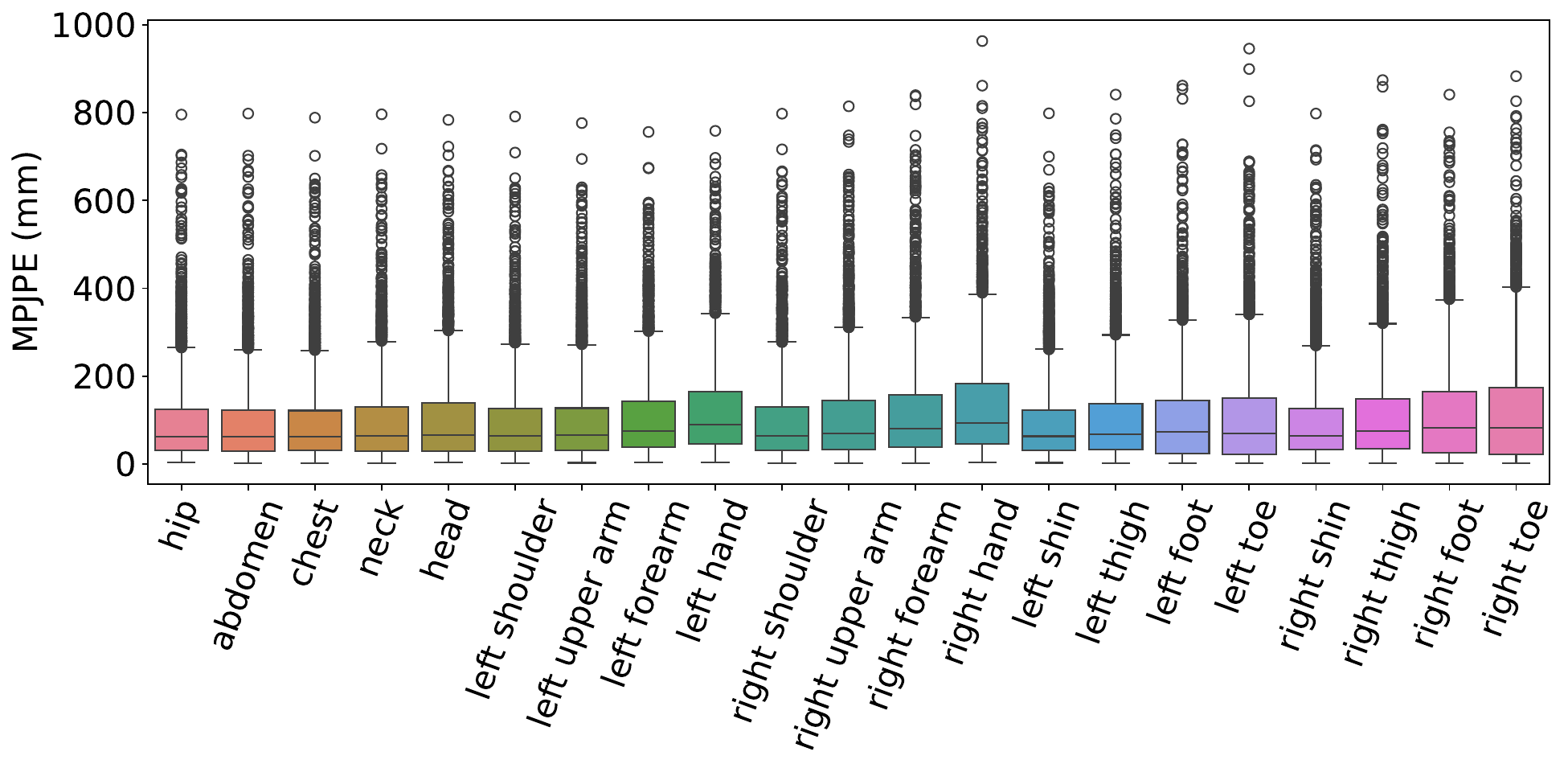}
    \caption{MPJPE for each joint using EqMotion with GT as input and a forecasting horizon of 1000 ms.}
    \label{fig:error-per-joint}
\end{figure}

3D-HPF in robot's perspective is the task of predicting the future pose of the human user with the sensors of the robot.
The pose at time $t$ can be denoted as $X_t\in\mathbb R^{D\times J_h}$, where $D=3$ is the dimension of the space and $J_h=21$ is the total number of joints in a human skeleton ($X_t$ is the set of all joint positions in 3D). Correspondingly, 3D-HPF means predicting $X_{t+1:t+K}$ based on $X_{t-T+1:t}$, where $X_{i:j}=X_i,X_{i+1},\ldots,X_j$, and $K$ is the \emph{horizon}.
In line with widely-used experimental protocols~\cite{h36m_pami,sofianos2021space}, we set $T=10$ 
and $K=4$ (roughly 400 ms) or $K=10$ (roughly 1000 ms), two cases referred to as \emph{short-term} and \emph{long-term} forecasting, respectively. We used average MPJPE over the $K$ predicted frames (average MPJPE) or MPJPE over the $K^{th}$ predicted frame (final MPJPE) as performance metrics.

The pose forecasting baselines we applied are STS-GCN~\cite{sofianos2021space}, SiMLPe~\cite{guo2023back} and EqMotion~\cite{xu2023eqmotion}. All three trained using MPJPE as a loss function without applying augmentation. The 
training was performed using the 21-joint poses obtained with the \optitrack sensor as a ground-truth.

Each baseline has three variants corresponding to different assumptions about the input data. The first variant, referred to as \emph{GT}, assumes that the robot can access all ground-truth joints in the human skeleton, the second (\emph{GT+R}) assumes that the robot can access only the joints visible to its sensors, the third (\emph{HRNet+D+R}) represents the 3D pose as shown in Section ~\ref{subsec:pose-estimation-eval}. GT+R deals with an input sequence of incomplete poses. These cannot be processed with the forecasting baselines above and, in general, with any of the approaches in the literature. Therefore, we used a diffusion-based time series imputation model, the CSDI~\cite{tashiro2021csdi}, built on a cascade of transformer blocks with skip connections. Such a model takes as input a sequence of incomplete poses and uses them to condition the generation of a complete pose, reconstructing the position of missing joints. The same applies to HRNet+D+R because the input poses can be incomplete for this variant too. 

Table~\ref{tab:pose-forecasting} shows the results for the three variants of every baseline. GT achieves the best results, while HRNet+D+R, corresponding to the most challenging task, shows the worst performance. EqMotion~\cite{xu2023eqmotion} is the baseline giving the best absolute results \emph{when in the presence of GT data}: 43 mm, on average, over the 400 ms horizon, and 70 mm over the 1000 ms horizon. However, STS-GCN~\cite{sofianos2021space} bridges the performance gap with EqMotion when the data is noisier like, \eg, in the HRNet+D+R case: the best average MPJPE is 313 mm over the 400 ms horizon and 332 mm over the 1000 ms horizon, while EqMotion achieves an MPJPE of 309 mm over the 400 ms horizon, and of 333 mm over the 1000 ms horizon.

Finally, we computed the MPJPE for each joint using the baseline with the smallest average error, \ie{}, EqMotion~\cite{xu2023eqmotion}, with GT as input (see Fig.~\ref{fig:error-per-joint}). We also estimated the correlation $r$ between these errors and the average velocity of each human joint with the Pearson coefficient ($r$=0.79, $p$=1.79e-05), noticing that the faster a joint moves, the harder it is to predict its trajectory in the future.

\subsection{Collision Prediction}
\label{sec:collision-detection}

\begin{figure*}[!ht]
    \centering
    \includegraphics[width=\linewidth]{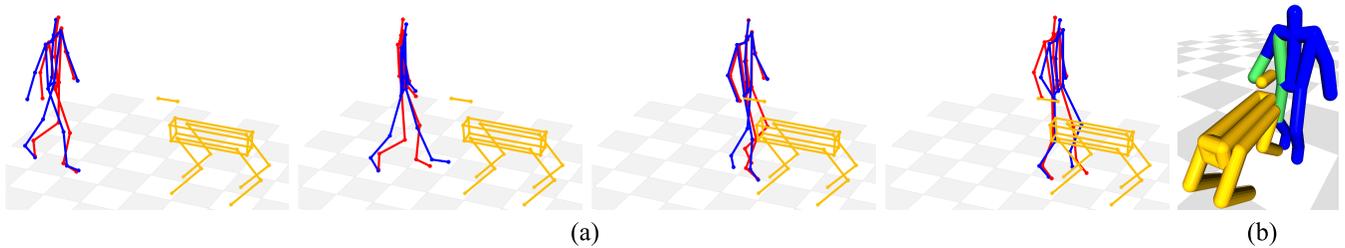}

    \caption{Qualitative results for the pose forecasting with the 1000 ms horizon. (a) shows the human pose forecasted in blue along with the ground truth in red. At the end of the sequence, an accidental collision occurs. In (b), the collision (highlighted in green) is detected as explained in Sec.~\ref{sec:collision-detection}. The forecasting approach used is EqMotion~\cite{xu2023eqmotion} on the GT data.}
    \label{fig:qualitative-forecasting} 
\end{figure*}

\begin{table*}[!htp]\centering
\caption{Performance of the different collision prediction methods with a 1000 ms horizon in terms of accuracy, sensitivity, and specificity score. The evaluation is divided into the four categories of contacts represented in the \ours dataset.}\label{tab:collision}
\scriptsize
\resizebox{\textwidth}{!}{%
\begin{tabular}{lc|rrr|rrr|rrr|rrr}\toprule
    \multirow{2}{*}{\textbf{Method}}&
    \multirow{2}{*}{\textbf{Input Type}}
    &\multicolumn{3}{c|}{\textbf{Unintended}} &\multicolumn{3}{c|}{\textbf{Touch}}
    &\multicolumn{3}{c|}{\textbf{Kick}}
    &\multicolumn{3}{c}{\textbf{Punch}}
    \\
    & 
    & \textbf{Acc.$\uparrow$} & \textbf{Sen.$\uparrow$} & \textbf{Spec.$\uparrow$} 
    & \textbf{Acc.$\uparrow$} & \textbf{Sen.$\uparrow$} & \textbf{Spec.$\uparrow$} 
    & \textbf{Acc.$\uparrow$} & \textbf{Sen.$\uparrow$} & \textbf{Spec.$\uparrow$} 
    & \textbf{Acc.$\uparrow$} & \textbf{Sen.$\uparrow$} & \textbf{Spec.$\uparrow$} 
    \\
\midrule

STS-GCN~\cite{sofianos2021space} & GT &0.91 & 0.91 & 0.91 & 0.94 & 0.91 & 0.99 & 0.75 & 0.46 & 0.96 & 0.74 & 0.78 & 0.70 \\
SiMLPe~\cite{guo2023back} & GT &0.93 & 0.93 & 0.92 & 0.95 & 0.93 & 0.98 & 0.81 & 0.83 & 0.79 & 0.79 & 0.89 & 0.65\\
EqMotion~\cite{xu2023eqmotion} & GT &0.95 & 0.95 & 0.94 & 0.97 & 0.96 & 0.97 & 0.93 & 0.91 & 0.94 & 0.82 & 0.87 & 0.76 \\

\midrule

Depth-based & D &0.49 & 0.25 & 0.90 & 0.53 & 0.33 & 0.87 & 0.71 & 0.39 & 0.94 & 0.60 & 0.61 & 0.59 \\
EqMotion~\cite{xu2023eqmotion} & HRNet+D+R & 0.76 & 0.65 & 0.91 & 0.92 & 0.90 & 0.95 & 0.72 & 0.52 & 0.85 & 0.70 & 0.73 & 0.65\\
\bottomrule
\end{tabular}%
}
\end{table*}

CP in robot's perspective is the task of predicting whether the user and robot will have physical contact, irrespective of whether it happens intentionally or not. 

We are particularly focused on the contacts or collisions caused by humans with the robot, due to the intricate challenges associated with predicting human movements, especially when partially visible.
Table~\ref{tab:actions} shows that \ours includes four types of physical contact (all acted to the best of the participants' abilities). Since they differ significantly in terms of energy and limbs involved, we addressed them as different cases in the experiments (see below). The CP process takes as input a sequence of human poses $X_{t:t+K}$ (see above for the notation) and a sequence of robot's poses $Y_{t:t+K}=(Y_{t+1},\ldots,Y_{t+K})$, where $Y_t\in\mathbb R^{D\times J_r}$ ($D=3$ and $J_r$ is the number of joints of the robot). The sequence $Y_{t:t+K}$ is assumed to be known because the robot plans its actions in advance. The goal of the process is to check whether $X_{t:t+K}$ and $Y_{t:t+K}$ contain a \emph{physical contact}, meaning that two cylinders of radius $r=5.0$ cm centered around the skeletal links of \spot and user are closer than a threshold $t_h=10.0$ cm (see Fig.~\ref{fig:qualitative-forecasting}b). As performance metrics we used accuracy, sensitivity, and specificity~\cite{heo2019collision}, where sensitivity is $\textrm{TP}/(\textrm{TP}+\textrm{FN})$ (it measures how effectively the system avoids False Positives), while specificity is $\textrm{TN}/(\textrm{TN}+\textrm{FP})$ (it measures how effectively the system predicts True Negatives). 

We started the CP-robot experiments by feeding the methods of Sec.~\ref{subsec:pose-forecasting-eval} with the \optitrack ground-truth data. This provided us with an upper bound of the performance and showed that punches and kicks are the contacts most difficult to predict (see Tab.~\ref{tab:actions}), probably due to the speed and energy involved. As a confirmation, the contact corresponding to the lowest speed and energy (touch), is the one leading to the best performance. After these initial tests, we replaced the ground-truth data with the pose forecasts output by EqMotion~\cite{xu2023eqmotion} in its HRNet+D+R variant, completed by the CSDI~\cite{tashiro2021csdi} diffusion process (see Section~\ref{subsec:pose-forecasting-eval}). Tab.~\ref{tab:actions} shows that the performances decrease, but not to a major extent.

Finally, we evaluated a straightforward baseline referred to as \emph{Depth-Based} in Tab.~\ref{tab:collision}, showing that CP-robot requires sophisticated approaches to be addressed. The baseline is a linear regression over the future $K$ depth frames given $T$ previous frames. This allowed us to test whether any points are predicted to get closer than $t_h$. For our experiments, we set $T$ and $K$ to the values used for the pose forecasting baselines, \ie, $T=10$, $K=10$ and $t_h=10$ cm. As expected, the performances are lower than in other cases. The only exception is the kick, in terms of accuracy and specificity, probably because the robot's cameras capture users' legs more easily than other parts of the body.
\section{Conclusions}\label{concl}
\label{sec:conclusions}

We presented \ours, the first dataset focused on how quadruped robots ``see'' their users. The data include 1) video and depth streams captured with the sensors of a \spot, and 2) skeleton representations of users and \spot in interaction captured with an \optitrack \mocap (the skeleton joint localization error is lower than 1 mm). The interaction scenarios were designed around specific problems (see Section~\ref{evaluation}). However, the data enables one to address a much wider spectrum of problems, including, \eg, proxemic behavior~\cite{Mumm2011} and action recognition~\cite{Chrungoo2014} (the list is not exhaustive).

In all cases, the key-novelty is that the \spot sensors can capture only part of the users' body. This leaves open the challenging problem of reconstructing the full 3D skeleton of the users while having at disposition only a partial 2D image of them. To the best of our knowledge, this is the first corpus revolving around such a problem and, therefore, we enriched the data with benchmarks including reproducible protocols and baseline approaches. In this way, the experiments we presented can be replicated and the results of future works can be rigorously compared with those of this paper.

\bibliographystyle{IEEEtran}
\bibliography{bibb}

\end{document}